\documentclass[conference]{IEEEtran}
\IEEEoverridecommandlockouts
\usepackage{cite}
\usepackage{amsmath,amssymb,amsfonts}
\usepackage{algorithmic}
\usepackage{graphicx}
\usepackage{textcomp}
\usepackage{xcolor}
\def\BibTeX{{\rm B\kern-.05em{\sc i\kern-.025em b}\kern-.08em
    T\kern-.1667em\lower.7ex\hbox{E}\kern-.125emX}}

\ifCLASSOPTIONcompsoc
\usepackage[caption=false,font=normalsize,labelfon
t=sf,textfont=sf]{subfig}
\else
\usepackage[caption=false,font=footnotesize]{subfig}
\fi

\begin{document}

\title{Conditional Neural Processes for Self-Supervised Goal Discovery in Action Imitation
\thanks{This research has been conducted in the framework of a Starting Grant from the European Research Council (ERC) under the European Union’s Horizon 2020 research and innovation programme. G.A. No 804388, wHiSPER. This work was supported in part by JST CREST “Cognitive Mirroring” under Grant JPMJCR16E2;}
}

\author{
    \IEEEauthorblockN{Marco Gabriele Fedozzi\IEEEauthorrefmark{1}\IEEEauthorrefmark{2}, Yukie Nagai\IEEEauthorrefmark{3}, Francesco Rea\IEEEauthorrefmark{2}, Alessandra Sciutti\IEEEauthorrefmark{2}}
    \IEEEauthorblockA{
        \IEEEauthorrefmark{1} DIBRIS Department, University of Genoa}
    \IEEEauthorblockA{
        \IEEEauthorrefmark{2} CONTACT Unit, Italian Institute of Technology \\ \{marco.fedozzi, francesco.rea, alessandra.sciutti\}@iit.it
    }
    \IEEEauthorblockA{
        \IEEEauthorrefmark{3} International Research Center for Neurointelligence, The University of Tokyo \\ nagai.yukie@mail.u-tokyo.ac.jp 
    }
}

\maketitle

\begin{abstract}

Inspired by infants' ability to learn novel tasks by imitating others, in this work we evaluate to what extent Conditional Neural Processes can autonomously learn representations of actions that are informative about their underlying goal. We show that the addition of an attention mechanism in the hidden space fosters the emergence of robust goal representations early on in the observation of an action sequence, making it a promising model for a robotic self-supervised general learner.


\end{abstract}


\section{Introduction}

The Mirror Neuron System (MNS) has a central role in our ability to predict the evolution of observed actions \cite{bonini2017extended}, being therefore associated with the inference of others' mental states (Theory of Mind \cite{baron1997build}). The predictive ability of the MNS appears to work at multiple layers of abstraction, ranging from the evolution of the motor plan to its underlying \textit{goal} \cite{oztop2013mirror}.
The emergence of this forecasting ability during infancy is accredited to a "passive" observation of others and an "active" association of self-actions with the generated sensory feedback \cite{hunnius2014you, giudice2009programmed}. Both modes are ultimately connected to infants' ability to imitate others, providing a shortcut to associate drives, goals, and motions, without explicit rewards \cite{meltzoff2018human}.
This association requires the formation of distinct (but connected) representations of different actions in the absence of direct supervision.
This capability inspires cognitive robotics' solutions, as a good self-supervised representation learning can ease later learning of novel tasks\footnote{E.g. refer to BERT \cite{devlin2018bert} and its robustness to multiple fine-tuned tasks.}.
This work evaluates to which extent Deep Modality Blending Networks (DMBN) \cite{seker2022imitation}, a multimodal generative robotic control model inspired by the MNS, manifest those emergent properties. 

\begin{figure*}[t]
  \centering
  \includegraphics[width=0.5\linewidth]{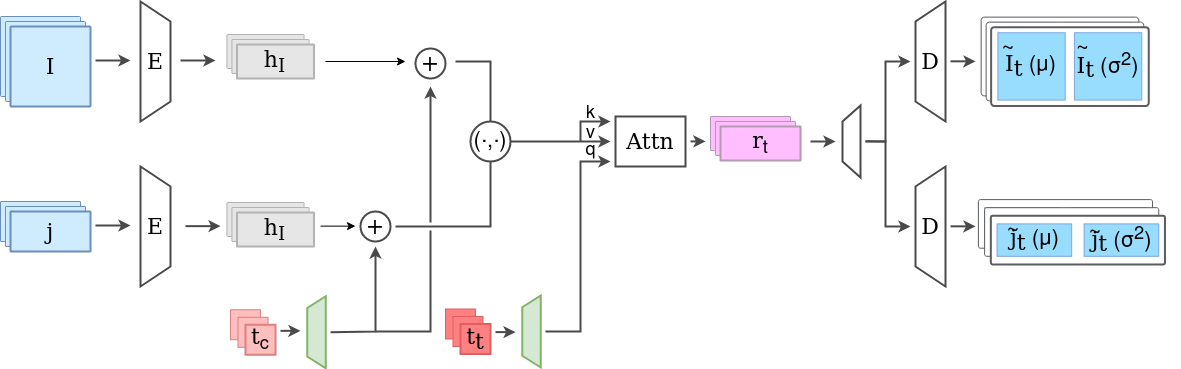}
  \caption{AttnDMBN architecture. The same color convention as in \cite{kim2019attentive} has been adopted. Networks with the same color share weights.}
  \label{fig:architecture}
\end{figure*}

\begin{figure*}[!h]
  \centering
      \subfloat[]{\includegraphics[width=0.24\linewidth]{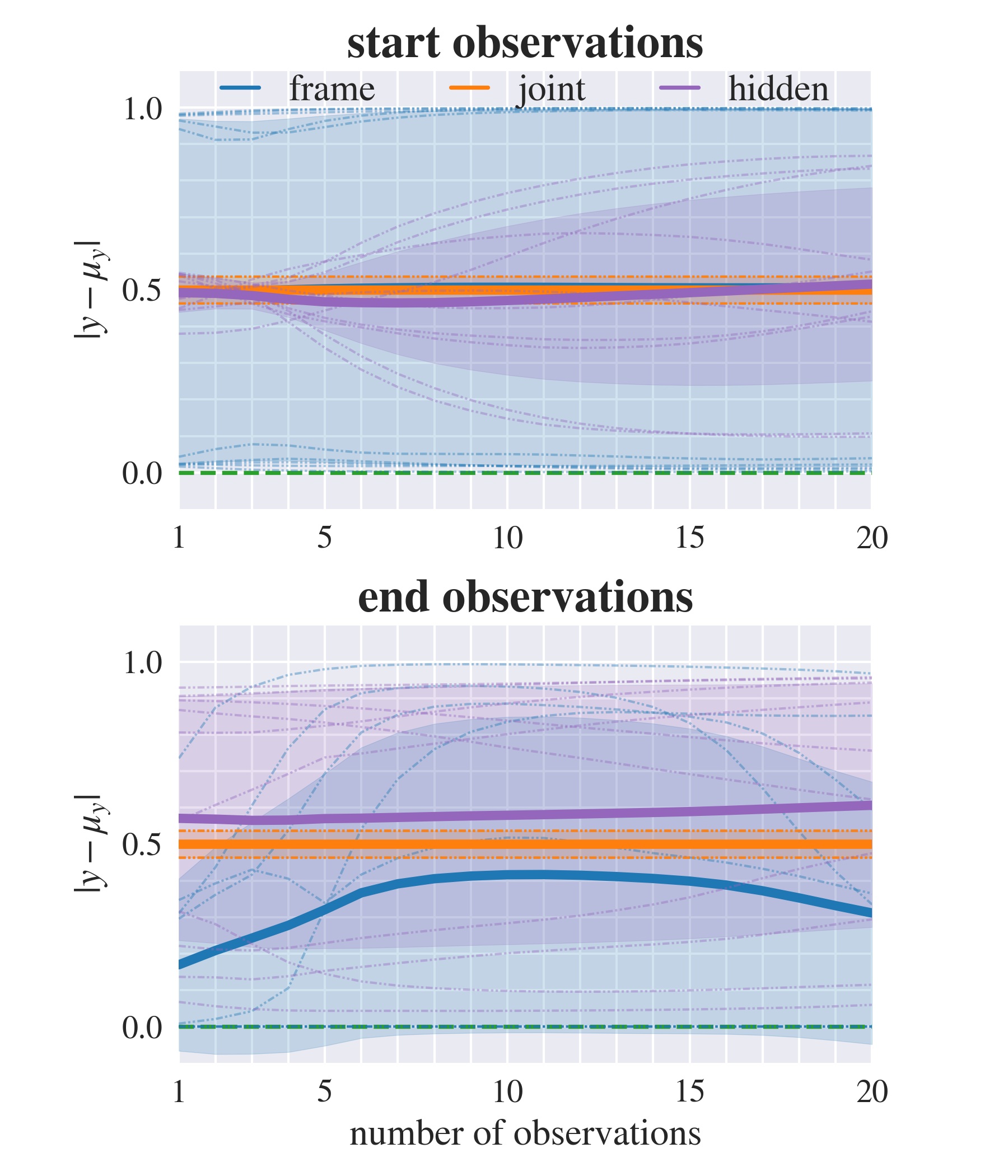}
      \label{fig:ae-out}}
      \hfill
      \subfloat[]{\includegraphics[width=0.24\linewidth]{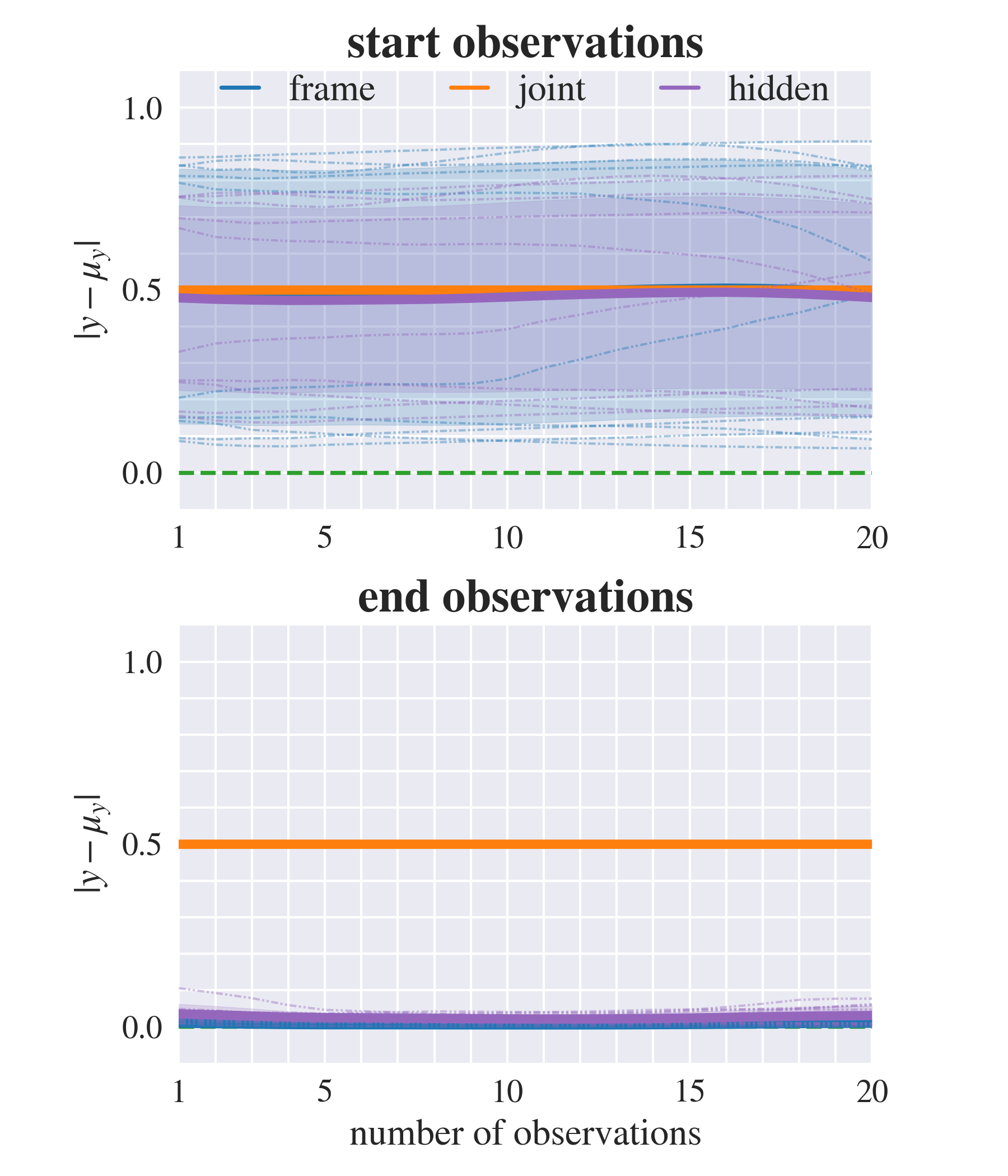}
      \label{fig:dmbn-out}}
      \hfill
      \subfloat[]{\includegraphics[width=0.24\linewidth]{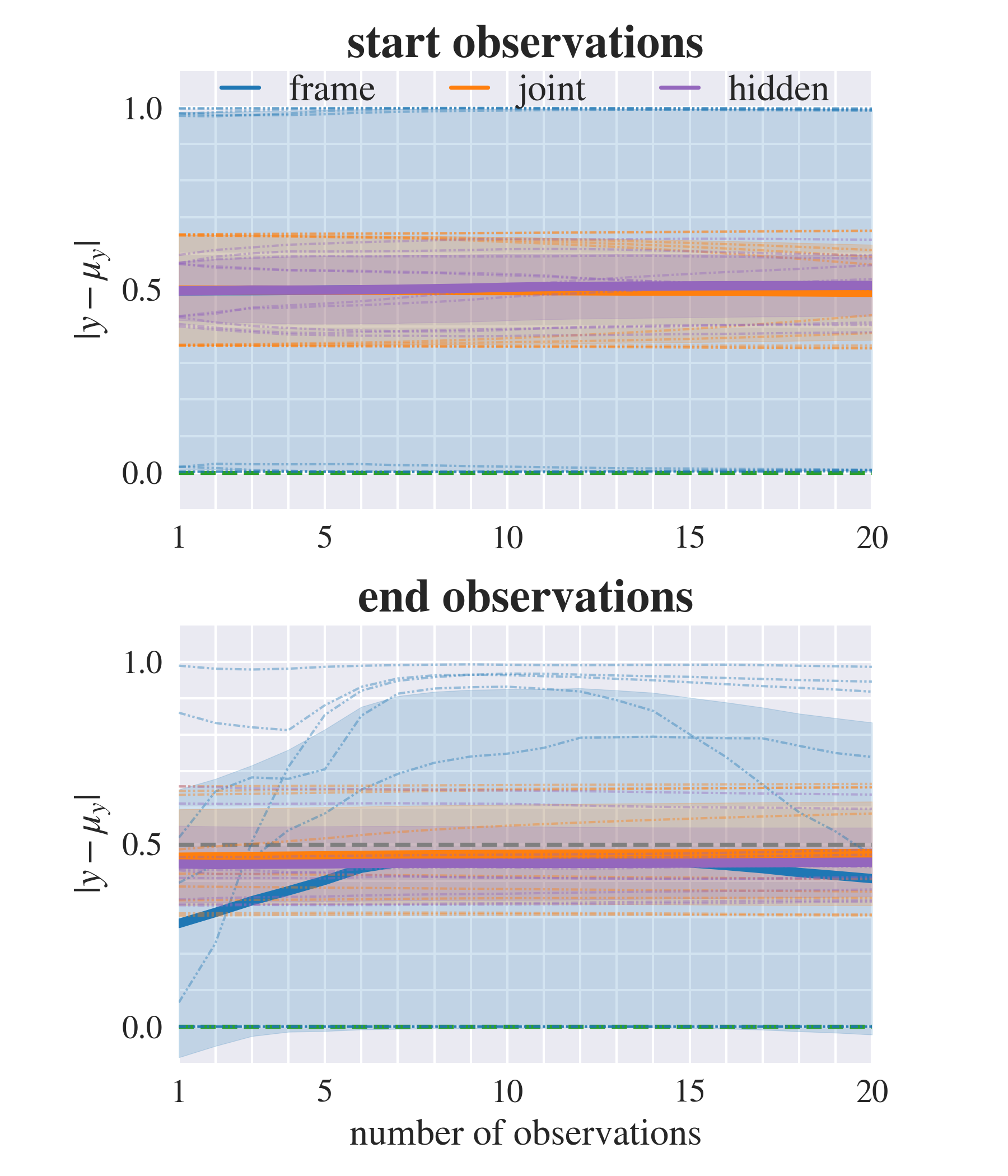}
      \label{fig:dmbn-pte-out}}
      \hfill
      \subfloat[]{\includegraphics[width=0.24\linewidth]{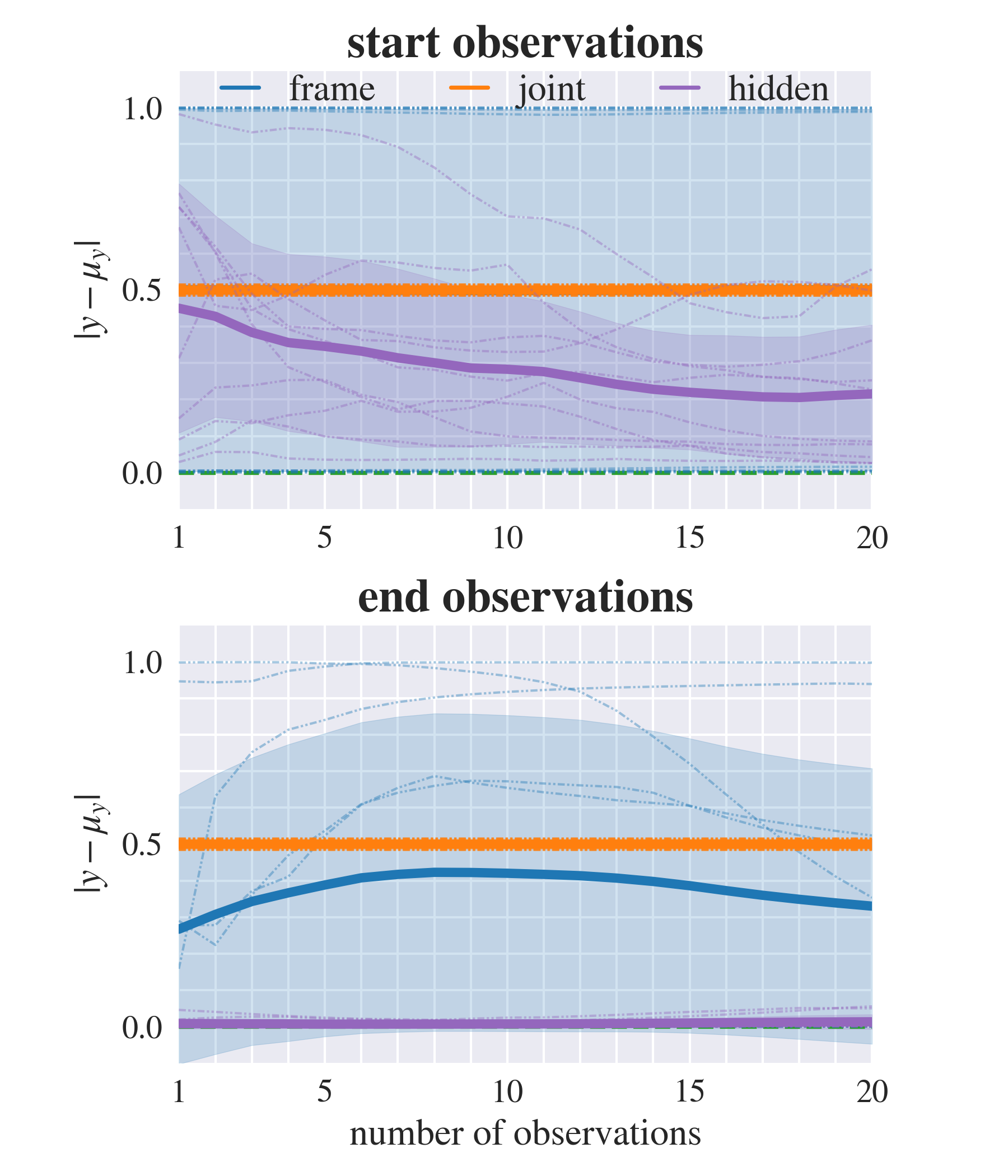}
      \label{fig:dmbn-attn-out}}
      
  \caption{Absolute distance of the mean classification from the correct label as function of the number of observations, either form the start or the end of the sequence. Solid lines and shaded area represent the mean and standard deviation of the behavior across the test set. Dash-dotted lines show the output for the individual test sequences.
  The \textit{frame} and \textit{joint} lines are the result of heads appended to the respective backbone encoder, whilst the \textit{hidden} head was placed after the CNP layer.
  \ref{fig:ae-out}) Autoencoder;
  \ref{fig:dmbn-out}) original DMBN;
  \ref{fig:dmbn-pte-out}) DMBN with Positional Time Encodings;
  \ref{fig:dmbn-attn-out}) DMBN with AttnCNP hidden layer.}
  \label{fig:out}
\end{figure*}

\section{Architectures}

Three variations of the original DMBN have been compared to it, aiming to pin-point the most impactful components for the formation of a good action representation.

\paragraph{DMBN}
This generative deep neural network reconstructs a bimodal vision-proprioceptive signal at an arbitrary point in time, conditioned on a set of observed instants. It uses Conditional Neural Processes \cite{garnelo2018conditional} to encode the observed set into an average \textit{representative element} to which the queried time is added before decoding to generate the required instant.

\paragraph{Autoencoder}
A fully connected net (FCN) replaces the CNP while the training objective consisted in reconstructing the observation set, without prediction.

\paragraph{DMBN-PTE}
The Positional Time Encoding \cite{fedozzi2023dmbnpte} version of the architecture projects input and target times with a FCN and adds them to the encoded modalities, instead of inserting them as input channels.

\paragraph{AttnDMBN}
Uses attention and Positional Time Encoding to generate the \textit{representative element}, inspired by \cite{kim2019attentive} but omitting the latent variable in favor of generating the variance at the output, and using the encoded observations as both \textit{keys} and \textit{values}, see Fig. \ref{fig:architecture}.

\section{Methodology}

To assess DMBN action representation quality, classification heads were appended to the unimodal encoders and the hidden projection layer of frozen backbones, inspired by \cite{bakhtiari2021functional}. They learned to classify the partially-observed motion into either \textit{"move"} or \textit{"grasp"} action. The same dataset presented in the original paper \cite{seker2022imitation} was utilized for both pretraining of the DMBN backbone and the later fine-tuning of the classification head.
Each sequence contained first-person images and joint angles of a simulated robotic arm grasping or displacing an object.
Backbone training used an Observation Set with 1 to 5 random time instants, expanded with other 1 to 5 random couples to form the Target Set. The frozen encoder plus the head was instead trained with 10 random observations, and the action performed as the target label.
Test subsequences between 1 and 20 items long, drawn from either the start or the end of the action, were used to evaluate the representation quality of hidden layers and correlate it with the amount of observed information and its proximity to the motion end\footnote{The intent behind an observed motion could remain opaque until midway through it, e.g. a grasp and move with a similar approach trajectory. The end of a sequence instead contains unambiguously the information on the action performed, given the displacement of the object.}.
Applying a \textit{softargmax} to the two head logits yielded a Bernoulli distribution, informative of the confidence in classifying an observed sequence.

\section{Results}

Fig. \ref{fig:out} depicts the absolute distance between the mean of the derived Bernoulli and the ground truth label across the four different nets, as a function of the observed subsequence length and its placement in the full sequence. A distance of $1.0$ corresponds to systematically assigning the wrong label; a distance of $0.5$ reflects either high uncertainty (close values of the head logits) or constant output (that evens out given the balanced dataset); a distance of $0.0$ signifies perfect predictive ability.
From Fig. \ref{fig:ae-out} emerges that the AE overall fails to predict the action from the input, as all distance sequences are at chance level.
The original DMBN in Fig. \ref{fig:dmbn-out} learns a representation with a high content of goal information when the final instants are observed, that does not extend to long scale prediction when observing the motion beginning.
Inserting time as positional encodings further removes goal-related information content from the representations, as seen in Fig. \ref{fig:dmbn-pte-out}.
Said information is also absent in each unimodal encoding of the AttnDMBN, but emerges in the output of the head appended after the attention layer, Fig. \ref{fig:dmbn-attn-out}. 



\section{Discussion}

Comparing the AE Fig. \ref{fig:ae-out} with the DMBN Fig. \ref{fig:dmbn-out} we can confirm that prediction is a good task for learning strong representations, as reconstruction alone does not foster the encoding of goal information. Notice that DMBN is however incapable of predicting the action form the first instants. 
The poor classification from the unimodal encodings of DMBN-PTE \ref{fig:dmbn-pte-out} is aligned with expectations, given the lack of time information in the input (behaving somewhat closer to the AE). The failed goal alignment of the representation after the CNP is again unsurprising considering that it is formed by simple non-linear projections of the previous ones.
More interesting is the output of the AttnDMBN, Fig. \ref{fig:dmbn-attn-out}: action information is present in the post-attention representation. The descending error curve in the top graph shows that the net learned to infer the goal also from observations at the beginning of the action.


\section{Conclusions}

The evaluation of the different models confirmed that Neural Processes could form robust task-agnostic action representations in self-supervised frameworks.
If the vanilla CNP fails representing predictive goal information, this ability is instead achieved using attentive layers. This highlights the impact of the way in which time is queried, making AttnDMBN a promising model to build a MNS-inspired architecture.
% ...

\bibliography{refs}

\end{document}